\documentstyle[aclap]{article}

\newcommand{\ignore}[1]{}

\newcommand{\BW}{BalancedWinnow}
\newcommand{\BWp}{$\mbox{BalancedWinnow}^{+}$}
\newcommand{\PW}{PositiveWinnow}
\newcommand{\PT}{Perceptron}

\newcommand{\tc}{text categorization}  
\newcommand{\ra}{\rightarrow}

\newenvironment{mytable}{\begin{table*}}{\end{table*}}

\newdimen\digitwidth
\setbox0=\hbox{\rm0}
\digitwidth=\wd0

\title{\vspace{-0.5in}
Mistake-Driven Learning in Text Categorization} 
\author{{\bf Ido Dagan}\thanks{
		       Partly supported by a grant no. 8560195 from the Israeli 
		       Ministry of Science.}\\
			      {\small Dept. of Math. \& CS}\\
			      {\small Bar Ilan University} \\
			      {\small Ramat Gan 52900, Israel}\\
			      {\small {\tt dagan@cs.biu.ac.il} }
\And
        {\bf Yael Karov}                                              \\
			      {\small Dept. of Appl. Math. \& CS}\\
			      {\small Weizmann Institute of Science} \\
			      {\small Rehovot 76100, Israel}\\
			      {\small {\tt yaelk@wisdom.weizmann.ac.il} }
\And 
        {\bf Dan Roth}\thanks{
                       Partly supported by a grant from the Israeli Ministry of Science.
		       Part of this work was done while visiting at Harvard University,
		       supported by ONR grant N00014-96-1-0550.}\\
			      {\small Dept. of Appl. Math. \& CS}\\
			      {\small Weizmann Institute of Science} \\
			      {\small Rehovot 76100, Israel}\\
			      {\small {\tt danr@wisdom.weizmann.ac.il} }
}
\input{psfig}
\begin{document}

\maketitle
\vspace{-0.5in}

\begin{abstract}

Learning problems in the text processing domain often map the text to
a space whose dimensions are the measured features of the text, e.g.,
its words. Three characteristic properties of this domain are (a) very
high dimensionality, (b) both the learned concepts and the instances
reside very sparsely in the feature space, and (c) a high variation in
the number of active features in an instance. In this work we study
three mistake-driven learning algorithms for a typical task of this
nature -- text categorization. 

We argue that these algorithms --
which categorize documents by learning a linear separator in the feature space --
have a few properties that make them ideal for this domain. We then 
show that a quantum leap in performance is achieved when we further 
modify the algorithms to better address some of the specific 
characteristics of the domain.  In particular, we
demonstrate (1) how variation in document length can be tolerated by
either normalizing feature weights or by using negative
weights, (2) the positive effect of applying a threshold range in
training, (3) alternatives in considering feature frequency, and (4)
the benefits of discarding features while training.  

Overall, we present an algorithm, a variation of Littlestone's Winnow,
which performs significantly better than any other algorithm
tested on this task using a similar feature set.

\end{abstract}

\section{Introduction} \label{sec:intro}

Learning problems in the natural language and text processing domains
are often studied by mapping the text to a space whose dimensions are 
the measured features of the text, e.g., the words appearing in a document. 
Three characteristic properties of this domain are 
(a) very high dimensionality, 
(b) both the learned concepts and the instances reside very sparsely in
the feature space and, consequently, 
(c) there is a high variation in the number of active features in an instance.

Multiplicative weight-updating algorithms such as Winnow \cite{Littlestone88} 
have been studied extensively in the theoretical learning literature.
Theoretical analysis has shown that they have exceptionally good behavior in 
domains with these characteristics, and in particular in the presence of 
irrelevant attributes, noise, and even a target function changing in time 
\cite{Littlestone88,LittlestoneWa94,HerbsterWa95}, 
but only recently have people started to use them in applications
\cite{GoldingRo96,Lewisetal96,CohenSi96}.
We address these claims empirically in an important application
domain for machine learning -- text categorization.
In particular, we study mistake-driven learning algorithms that are based
on the Winnow family, and investigate ways to apply them in domains with 
the above characteristics. 

The learning algorithms studied here offer a large space of choices to 
be made and, correspondingly, may vary widely in performance when applied 
in specific domains. 
We concentrate here on the text processing domain, with the characteristics
mentioned above, and explore this space of choices in it.

In particular, we investigate three variations of on-line prediction 
algorithms and evaluate them experimentally on large text categorization
problems. 
The algorithms we study are all learning algorithms for linear functions.
They are used to categorize documents by learning, for each category,  a 
linear separator in the feature space. 
The algorithms differ by whether they allow the use of negative or only 
positive weights and by the way they update their weights during
the training phase.

We find that while a vanilla version of these algorithms performs
rather well, a quantum leap in performance is achieved when we modify 
the algorithms to better address some of the specific characteristics 
we identify in textual domains.
In particular, we address problems such as wide variations in document sizes, 
word repetitions and the need to rank documents rather than just decide 
whether they belong to a category or not.
In some cases we adopt solutions that are well known in the IR literature
to the class of algorithms we use; in others we modify known algorithms
to better suit the characteristics of the domain. 
We motivate the modifications to the basic algorithms and justify them 
experimentally by exhibiting their contribution to improvement in
performance. 
Overall, the best variation we investigate, performs significantly
better than any known algorithm tested on this task, using a similar set
of features.

The rest of the paper is organized as follows:
The next section describes the task of text categorization, how we model
it as a classification task, and some related work.
The family of algorithms we use is introduced in
Section~\ref{sec:algs} and the extensions to the basic algorithms,  
along with their experimental evaluations, is presented in 
Section~\ref{sec:extend}.
In Section~\ref{sec:exp} we present our final experimental results 
and compare them to previous works in the literature.

\section{Text Categorization} \label{sec:text}

In \tc, given a text document and a collection of potential classes, 
the algorithm decides which classes it belongs to, or how strongly 
it belongs to each class.
For example, possible classes (categories) may be
$\{ {\it bond}\}, \{{\it loan}\}, \{{\it interest}\}, 
\{{\it acquisition } \}$.
Documents that have been categorized by humans are usually used as 
training data for a \tc~ system; later on, the trained system is used 
to categorize new documents.
Algorithms used to train \tc~ systems in information retrieval (IR) are 
often ad-hoc and poorly understood. In particular, very little is known
about their generalization performance, that is, their behavior on 
documents outside the training data.
Only recently, some machine learning techniques for training linear 
classifiers have been used and shown to be effective in this domain 
\cite{Lewisetal96,CohenSi96}. These techniques have the advantage that 
they are better understood from a theoretical standpoint, leading to 
performance guarantees and guidance in parameter settings. 
Continuing this line of research we present different algorithms and focus
on adjusting them to the unique characteristics of the domain, yielding
good performance on the categorization task.

\subsection{Training Text Classifiers } \label{sec:training}

Text classifiers represent a document as a set of features 
$d = \{f_1,f_2, \ldots f_m\}$, where $m$ is the number of {\em active} 
features in the document, that is, features that occur in the document.
A feature $f_i$ may typically represent a word $w$, a set $
{\em w_1, \ldots w_k}$ of words \cite{CohenSi96} or a phrasal structure 
\cite{Lewis92,TzerasHa93}.
The {\em strength} of the feature $f$ in the document $d$ is denoted by
$s(f,d)$. The strength is usually a function of the number of times $f$ 
appears in $d$ (denoted by $n(f,d)$). The strength may be used 
only to indicate the presence or absence of $f$ in the document, in
which case it takes on only the values $0$ or $1$, it may be equal to 
$n(f,d)$, or it can take other values to reflect also the size of the document.

In order to rank documents, for each category, a \tc~ system keeps a function
$F_c$ which, when evaluated on $d$, produces a score $F_c(d)$.  A decision is
then made by assigning to the category $c$ only those documents that exceed 
some threshold, or just by placing at the top of the ranking documents with 
the highest such score.

A {\em linear} text classifier represents a category as a weight vector  
$w_c = ( w(f_1,c), w(f_2,c), \ldots w(f_n,c) ) \equiv (w_1, w_2,
\ldots w_n)$, where $n$ is the total number of features in the domain and 
$w(f,c)$ is the weight of the feature $f$ for this category. It evaluates 
the score of the document by computing the dot product: 
$$ F_c(d) = \sum_{f \in d} s(f,d) \cdot w(f,c).$$

The problem is modeled as a supervised learning problem. The algorithms
use the training data, where each document is labeled by zero or more 
categories, to learn a classifier which classifies new texts.
A document is considered as a positive example for all categories with
which it is labeled, and as a negative example to all others.  

The task of a training algorithm for a linear text classifier is to find a 
weight vector which best classifies new text documents. 
While a linear text classifier is a linear separator in the space defined 
by the features, it may not be linear with respect to the document, if one 
chooses to use complex features such as conjunctions of simple features.  
In addition, a training algorithm may give also advice on the issue of 
feature selection, by reducing the weight of non-important features and 
thus effectively discarding them.

\subsection{Related Work} \label{sec:related}
Many of the techniques previously used in \tc~ make use of linear classifiers,
mainly for reasons of efficiency. 
The classical vector space model, which ranks documents using a
nonlinear similarity measure (the ``cosine correlation'') \cite{SaltonBu83}
can also be recast as a linear classification by incorporating length 
normalization into the weight vector and the document vector features values.
State of the art IR systems determine the strength of a term based on three
values: 
(1) the frequency of the feature in the document ({\em tf}), 
(2) an inverse measure of the frequency of the feature throughout 
the data set ({\em idf}),
and (3) a normalization factor that takes into account the length of the 
document. 
In Sections~\ref{sec:length} and \ref{sec:repeat} we discuss how we incorporate
those ideas in our setting.

Most relevant to our work are non-parametric methods, which seem to yield 
better results than parametric techniques. 
Rocchio's algorithm \cite{Rocchio71}, one of the most commonly used 
techniques, is a batch method that works in a relevance feedback context.
Typically, classifiers produced by the Rocchio algorithm are restricted to 
having nonnegative weights.
An important distinction between most of the classical non-parametric methods
and the learning techniques we study here is that in the former case, there
was no theoretical work that addressed the generalization ability of the learned 
classifier, that is, how it behaves on new data.

The methods that are most similar to our techniques are the on-line algorithms
used in \cite{Lewisetal96} and \cite{CohenSi96}.
In the first, two algorithms, a multiplicative update and additive update 
algorithms suggested in \cite{KivinenWa95} are evaluated in the \tc~ domain, 
and are shown to perform somewhat better than Rocchio's algorithm. 
While both these works make use of multiplicative update algorithms, as
we do, there are two major differences between those studies and the current
one. First, there are some important technical differences between the 
algorithms used. Second, the algorithms we study here are mistake-driven; 
they update the weight vector only when a mistake is made, and not after
every example seen. The Experts algorithm studied in \cite{CohenSi96} is
very similar to a basic version of the \BW~ algorithm which we study here. 
The way we treat the negative weights is different, though, and  significantly 
more efficient, especially in sparse domains (see Section~\ref{sec:algorithms}). 
Cohen and Singer experiment also, using the same algorithm, 
with more complex features (sparse n-grams) and show that, as expected,
it yields better results.  

Our additive update algorithm, \PT, is somewhat similar to what is used in 
\cite{Wieneretal95}. They use a more complex representation, a multi-layer 
network, but this additional expressiveness seems to make training more
complicated, without contributing to better results.

\subsection{Methodology} \label{sec:methodology}

We evaluate our algorithms on the the Reuters-22173 text collection 
\cite{Lewis92}, one of the most commonly used benchmarks in the literature. 

For the experiments reported 
In Sections~\ref{sec:exp1} we explore and compare different variations of the 
algorithms; we evaluate those on two disjoint pairs of a training set and a 
test set, both subsets of the Reuters collection. Each pair consists of 2000 
training documents and 1000 test documents, and was used to train and test the
classifier on a sample of 10 topical categories. 
The figures reported are
the average results on the two test sets. 

In addition, we have tested our final version of the classifier on two
common partitions of the complete Reuters collection, and compare the
results with those of other works. 
The two partitions used are those of 
Lewis \cite{Lewis92} (14704 documents for training, 6746 for testing) and 
Apte \cite{Apte-et-al94} (10645 training, 3672 testing, omitting
documents with no topical category). 

To evaluate performance, the usual measures of recall and precision were used.
Specifically, we measured the effectiveness of the classification by 
keeping track of the following four numbers:
\begin{itemize}

\item
$p_1$ = number of correctly classified class members
\item
$p_2$ = number of mis-classified class members
\item
$n_1$ = number of correctly classified non-class members
\item
$n_2$ = number of mis-classified non-class members

\end{itemize}
In those terms, the {\em recall} measure is defines as $p_1/p_1 + p_2$,
and the {\em precision  } is defined as $p_1/p_1 + n_2$.
Performance was further summarized by a {\em break-even point} -- a 
hypothetical point, obtained by interpolation, in which precision equals 
recall.

\section{On-Line learning algorithms} \label{sec:algs}

In this section we present the basic versions of the learning 
algorithms we use. The algorithms are used to learn a classifier $F_c$ for 
each category $c$. 
These algorithms use the training data, where each document is labeled by
zero or more categories, to learn a weight vector which is used later on,
in the test phase, to classify new text documents.
A document is considered as a positive example for all categories with which 
it is labeled, and as a negative example to all others.
The algorithms are on-line and mistake-driven.
In the on-line learning model, learning takes place in a sequence of trials. 
On each trial, the learner first makes a prediction and then receives 
feedback which may be used to update the current hypothesis (the vector
of weights).
A mistake-driven algorithm updates its hypothesis only when a mistake is made. 
In the training phase, given a collection of examples, we may repeat this 
process a few times, by iterating on the data. In the testing phase, the same
process is repeated on the test collection, only that the hypothesis is not updated.
 
Let $n$ be the number of features of the current category.
For the remainder of this section we denote
a training document with $m$ active features by 
$d = (s_{i_1}, s_{i_2}, \ldots s_{i_m})$, where $s_{i_j}$ stands for the strength
of the $i_j$ feature.  The label of the document
is denoted by $y$; $y$ takes the value $1$ if the document is relevant to the
category and $0$ otherwise.
Notice, that we care only about the active features in the domain,
following \cite{Blum92}. 
The algorithms have three parameters: a threshold $\theta$, and two update 
parameters, a {\it promotion\/} parameter $\alpha$ and a {\it demotion\/} 
parameter $\beta$.

\subsubsection*{Positive Winnow \cite{Littlestone88}:}
The algorithm keeps an $n$-dimensional weight vector 
$w = (w_1, w_2, \ldots w_n)$, $w_i$ being the weight of the $i$th feature,
which it updates whenever a mistake is made. 
Initially, the weight vector is typically set to assign equal positive weight
to all features. (We use the value $\theta/d$, where $d$ is the average number
of active features in a document; in this way initial scores are close to $\theta$.)
The promotion parameter is $\alpha>1$ and the demotion is $0<\beta<1$.

For a given instance $(s_{i_1}, s_{i_2} \ldots ,s_{i_m})$ the algorithm 
predicts $1$ iff 
$$\sum_{j=1}^{m} w_{i_j} s_{i_j} > \theta,$$ 
where $w_{i_j}$ is the weight corresponding to the active feature indexed by
${i_j}$. The algorithm updates its hypothesis only when a mistake is made, 
as follows:
(1) If the algorithm predicts $0$ and the label is $1$ (positive example)
then the weights of all the active features are promoted ---
the weight $w_{i_j}$ is multiplied by $\alpha$.
(2) If the algorithm predicts $1$ and the received label is $0$ (negative
example) then the weights of all the active features are demoted ---
the weight $w_{i_j}$ is multiplied by $\beta$. 
In both cases, weights of inactive features maintain the same value.

\subsubsection*{Perceptron \cite{Rosenblatt58}}
As in \PW, in \PT~ we also keep an $n$-dimensional weight vector 
$w = (w_1, w_2, \ldots w_n)$ whose entries correspond to the set 
of potential features, which is updated whenever a mistake is made.
As above, the initial weight vector is typically set to assign equal weight to
all features. 
The only difference between the algorithms is that in this case the weights 
are updated in an {\em additive} fashion. 
A single update parameter  $\alpha > 0$ is used, and a weight 
is promoted by {\em adding} $\alpha$ to its previous value, and is demoted by 
{\em subtracting } $\alpha$ from it.
In both cases, all other weights maintain the same value.

\subsubsection*{Balanced Winnow \cite{Littlestone88}:}
In this case, the algorithm keeps two weights, $w^{+},w^{-}$, for each
feature. The overall weight of a feature is the difference between these 
two weights, thus allowing for negative weights. For a given instance 
$(s_{i_1}, s_{i_2} \ldots ,s_{i_m})$ the algorithm predicts $1$ iff 
\begin{equation}\label{eq:balanced}
\sum_{j=1}^{m} (w^{+}_{i_j} - w^{-}_{i_j})  s_{i_j} > \theta,
\end{equation}
where $w^{+}_{i_j},w^{-}_{i_j}$ correspond to the active feature indexed 
by ${i_j}$.
In our implementation,  the weights $w^{+}$ are initialized to  $2\theta/d$
and the weights $w^{-}$ are set to $\theta/d$, where $d$ is the
average number of active features in a document in the collection.

The algorithm updates the weights of active features only when a
mistake is made, as follows: 
(1) In the promotion step, following a mistake on a positive example, 
the positive part of the weight is promoted, 
$w^{+}_{i_j} \leftarrow \alpha \cdot w^{+}_{i_j}$
while the negative part of the weight is demoted,
$w^{-}_{i_j} \leftarrow \beta \cdot w^{-}_{i_j}$. 
Overall, the coefficient of $s_{i_j}$ in Eq.~\ref{eq:balanced} increases 
after a promotion.
(2) In the demotion step, following a mistake on a negative example, 
the coefficient of $s_{i_j}$ in Eq.~\ref{eq:balanced} is decreased:
the positive part of the weight is demoted, 
$w^{+}_{i_j} \leftarrow \beta \cdot w^{+}_{i_j}$
while the negative part of the weight is promoted,
$w^{-}_{i_j} \leftarrow \alpha \cdot w^{-}_{i_j}$. 
In both cases, all other weights maintain the same value.

In this algorithm (see in Eq.~\ref{eq:balanced}) the coefficient of the $i$th 
feature can take negative values, unlike the representation used in \PW. 
There are other versions of the Winnow algorithm that allow the use of 
negative features: (1) Littlestone, when introducing the Balanced version, 
introduced also a simpler version -- a version of \PW~ 
with a duplication of the number of features.
(2) A version of the Winnow algorithm with negative features is used in 
\cite{CohenSi96}.
In both cases, however, whenever there is a need to update the weights, 
{\em all} the weights are being updated (actually, $n$ out of the $2n$). 
In the version we use, only weights of {\em active} features are being 
updated; this gives a significant computational advantage  when working in 
a sparse high dimensional space.

\subsection{Properties of the Algorithms} \label{sec:algorithms}

Winnow and its variations were introduced in Littlestone's seminal paper 
\cite{Littlestone88}; the theoretical behavior of multiplicative 
weight-updating algorithms for learning linear functions has been studied 
since then extensively.
In particular, Winnow has been shown to learn efficiently any linear 
threshold function \cite{Littlestone88}.
These are functions $F: \{0,1\}^{n} \ra \{0,1\}$ for which there exist 
real weights $w_1, \ldots, w_n$ and a real threshold $\theta$ such that
$F(s_1,\ldots,s_n) = 1$ iff $\sum_{i=1}^{n} w_i s_i \geq \theta$.
In particular, these functions include Boolean disjunctions and conjunctions
on $k \leq n$ variables and $r$-of-$k$ threshold functions
($1 \leq r \leq k \leq n$).
While Winnow is guaranteed to find a perfect separator if one exists, it also
appears to be fairly successful when there is no perfect separator.
The algorithm makes no independence or any other assumptions on the features, 
in contrast to other parametric estimation techniques (typically, Bayesian 
predictors) which are commonly used in statistical NLP.

Theoretical analysis has shown that the algorithm has exceptionally 
good behavior in the presence of irrelevant features, noise, and even a 
target function changing in time 
\cite{Littlestone88,Littlestone91,LittlestoneWa94,HerbsterWa95}, 
and there is already some empirical support for these claims 
\cite{Littlestone95,GoldingRo96,Blum95}.
The key feature of Winnow is that its mistake bound grows linearly with the 
number of {\em relevant\/} features and only logarithmically with the total 
number of features.
A second important property is being mistake driven. Intuitively, this makes 
the algorithm more sensitive to the relationships among the features --- 
relationships that may go unnoticed by an algorithm that is based on counts 
accumulated separately for each attribute.
This is crucial in the analysis of the algorithm 
as well as empirically \cite{Littlestone95,GoldingRo96}.

The discussion above holds for both versions of Winnow studied here, 
\PW~ and \BW. The theoretical results differ only slightly in the mistake 
bounds, but have the same flavor. However, the major difference between the 
two algorithms, one using only positive weights and the other allowing also 
negative weights, plays a significant role when applied in the current domain,
as discussed in Section~\ref{sec:extend}.

Winnow is closely related, and has served as the motivation for 
a collection of recent works on combining the ``advice" of different
``experts"\cite{LittlestoneWa94,CBFHHSW93,CBFHW94}. 
The features used are the ``experts" and the learning algorithm can be 
viewed as an algorithm that learns how to combine the classifications of 
the different experts in an optimal way.

The additive-update algorithm that we evaluate here, the \PT, goes back to
\cite{Rosenblatt58}. 
While this algorithm is also known to learn the target 
linear function when it exists, the bounds given by the Perceptron convergence 
theorem \cite{DudaHa73} may be exponential in the optimal mistake bound, 
even for fairly simple functions \cite{KivinenWa95a}.
We refer to \cite{KivinenWa95} for a thorough analysis of multiplicative 
update algorithms versus additive update algorithms.
In particular, it is shown that the number of mistakes the additive and 
multiplicative update algorithms make, depend differently on the domain 
characteristics. Informally speaking, it is shown that the multiplicative 
update algorithms have advantages in high dimensional problems (i.e., when 
the number of features is large) and when the target weight vector is sparse
(i.e., contain many weights that are close to $0$). 
This explains the recent success in using these methods
on high dimensional problems \cite{GoldingRo96} and suggests that 
multiplicative-update algorithms might do well on IR applications, provided 
that a good set of features is selected. On the other hand, it is shown that
additive-update algorithms have advantages when the examples are sparse in 
the feature space, another typical characteristics of the IR domain, which 
motivates us to study experimentally an additive-update algorithm as well.

\subsection{Evaluating the Basic Versions} \label{sec:exp1}

We started by evaluating the basic versions of the three
algorithms. 
The features we use throughout the experiments are single words, at
the lemma level, for nouns and verbs only, with minimal frequency of 3
occurrences in the corpus. 
In the basic versions the strength of the feature is taken to indicate only
the presence or absence of $f$ in the document, that is, it is either $1$ or $0$.
The training algorithm was run iteratively on the training set, until no 
mistakes were made on the training collection or until some upper bound (50) 
on the number of iterations was reached. 
The results for the basic versions are shown in the first column of
Table~\ref{tab:our}. 

\begin{mytable}[tbh]
\begingroup\catcode`?=\active%
\def?{\kern\digitwidth}%
\begin{center}
\begin{tabular}{||l|c|c|c|c|c|c||}
\hline
Algorithm & \multicolumn{6}{c||}{Version} \\
  & Basic   & Norm   & $\theta$-range & Linear Freq. & Sqrt. Freq & Discard \\
\hline
\hline
\BW       & 64.87   & NA     & 69.66           & 72.11       & 71.56       & 73.2         \\
\hline
\PW       & 55.56   & 63.56  & 65.80           & 67.20       & 69.67       & 70.0     \\
\hline
\PT       & 65.91   & NA     & 63.05           & 66.72       & 68.29       & 70.8         \\
\hline
\end{tabular}
\end{center}
\caption{Recall/precision break-even point (in percentages) for different
versions of the algorithm. Each figure is an average result for two
pairs of training and testing sets, each containing 2000 training
documents and 1000 test documents.}
\label{tab:our}
\endgroup
\end{mytable}

\section{Extensions to the Basic algorithms} \label{sec:extend}

\subsection{Length Variation and Negative features} \label{sec:length}

Text documents vary widely in their length and a text classifier
needs to tolerate this variation. This issue is a potential problem 
for a linear classifier which scores a document by summing the weights 
of all its active features: a long document may have a better chance of 
exceeding the threshold merely by its length.

This problem has been identified earlier on and attracted a lot of work
in the classical work on IR \cite{SaltonBu83}, 
as we have indicated in Section~\ref{sec:related}.
The treatment described there addresses at the same time at least two different concerns:
length variation of documents and feature repetition. In this section we
consider the first of those, and discuss how it applies to the algorithms we investigate.
The second concern is discussed in Section~\ref{sec:repeat}.

Algorithms that allow the use of negative features, such as \BW~and \PT,
tolerate variation in the documents length naturally, and thus have 
a significant advantage in this respect. In these cases, it can be
expected that the cumulative contribution of the weights and, in particular,
those that are not indicative to the current category, does not count
towards exceeding the threshold, but rather averages out to $0$. Indeed, 
as we found out, no special normalization is required when using these
algorithms. Their significant advantage over the unnormalized version of \PW~ 
is readily seen in Table~\ref{tab:our}.

In addition, using negative weights gives the text classifier more
flexibility in capturing ``truly negative'' features, where the
presence of a feature is indicative for the {\em irrelevance}  
of the document to the category. However, we found that this phenomenon
only rarely occurs in \tc~and thus the main use of the negative features is 
to tolerate the length variation of the documents.

When using \PW, which uses only positive weights, we no longer have this 
advantage and we seek a modification that tolerates the variation in length.
As in the standard IR solution, we suggest to modify $s(f,d)$, the strength 
of the feature $f$ in $d$,  by using a quantity
that is normalized with respect to the document size.

Formally, we replace the strength $s(f,d)$ (which may be determined in
several ways according to feature frequency, as explained below) by a
{\em normalized strength},
$$s^{n}(f,d) = \frac{s(f,d)}{\sum_{f\in d}s(f,d)}.$$

In this case (which applies, as discussed above, only for \PW), we also change 
the initial weight vector and initialize all the weights to $\theta$. 

Using normalization gives an effect that is similar to the use of negative 
weights, but to a lesser degree. The reason is that it is used uniformly; 
in long documents, the number of indicative features does not increase
significantly, but their strength, nevertheless, is reduced proportionally 
to the total number of features in the document.
In the long version of the paper we present a more thorough analysis of this
issue.

The results presented in Table~\ref{tab:our} (second column) show the
significant improvements achieved in \PW~ performance, when
normalization is used. In all the results presented from this point
on, positive winnow is normalized. 

\subsection{Using Threshold range}

Training a linear text classifier is a search for a weight vector in the 
feature space. The search is for a linear separator that best separates 
documents that are relevant to the category from those that are not. 
In general, there is no guarantee that a weight vector of this sort exists, 
even in the training data, but a good selection of features make
this more likely. 
While the basic versions of our algorithms search for linear 
separators, we have modified those so that our search for a linear classifier 
is biased to look for ``thick" classifiers.
To understand this, consider, for the moment, the case in which all the data 
is perfectly linearly separable. Then there will generally be many linear 
classifiers that separate the training data we actually see.
Among these, it seems plausible that we have a better chance of doing well 
on the unseen test data if we choose a linear separator that separates the 
positive and negative training examples as ``widely'' as possible.
The idea of having a wide separation is less clear when there is no perfect 
separator, but we can still appeal to the basic intuition.

Using a ``thick'' separator is even more important when documents are ranked rather
than simply classified; that is, when the actual score 
produced by the classifier is used in the decision process.
The reason is that if $F_c(d)$ is the score produced by the classifier 
$F_c$ when evaluated on the document $d$ then, under some assumptions
on the dependencies among the features, the probability that the document 
$d$ is relevant to the category $c$ is given by
$Prob(d \in c) = \frac{1}{1 + e^{- F_c(d)} } $.
This function, known as the {\em sigmoid} function, ``flattens" the decision 
region in a way that only scores that are far apart from the threshold value
indicate that the decision is made with significant probability.

Formally, among those weight vectors we would like to choose the 
hyper-plane with the largest ``separating parameter",
where the separating parameter $\tau$ is defined as the largest value for 
which there exists a classifier $F_c$ (defined by a weight vector $w$)  
such  that for all positive examples $d$, 
$F_c(d) > \theta + \tau/2$ and for all negative $d$, $F_c(d) < \theta - \tau/2$.

In this implementation we do not try to find the optimal $\tau$ (as is done
in \cite{CortesVa95}, but rather determine it heuristically. 
In order to find a ``thick" separator, we modify, in all three algorithms, 
the update rule used during the training phase as follows: 
Rather than using a single threshold we use two separate thresholds, 
$\theta^{+}$ and $\theta^{-}$, such that $\theta^{+} -\theta^{-} = \tau$.
During training, we say that the algorithm predicts $0$ (and makes
a mistake, if the example is labeled positive) when the score it assigns an 
example is below $\theta^{-}$. 
Similarly, we  say that the algorithm predicts $1$ when the score 
exceeds $\theta^{+}$.
All examples with scores in the range  $[\theta^{-},\theta^{+}]$ are 
considered mistakes.
(Parameters used: $\theta^{-}$=0.9, $\theta^{+}$ = 1.1, $\theta$ = 1).

The results presented in the third column of Table~\ref{tab:our} show
the improvements obtained when the threshold range is used. 
In all the results presented from this point on, all the algorithms use
the threshold range modification.

\subsection{Feature Repetition} \label{sec:repeat}

Due to the bursty nature of term occurrence in documents, as well as the
variation in document length, a feature may occur
in a document more than once. It is therefore important to consider
the frequency of a feature when determining its strength.
On one hand, there are cases where a feature is more indicative to the
relevance of the document to a category when it appears several times
in a document. On the other hand, in any long document, there may be some 
random feature that is not significantly indicative to the current category 
although it repeats many times. 
While the weight of $f$ in the weight vector of the
category, $w(f,c)$, may be fairly small, its cumulative contribution
might be too large if we increase its strength, $s(f,d)$, in proportion to
its frequency in the document.

As mentioned in Section~\ref{sec:related}, the classical IR literature has addressed 
this problem using the {\em tf} and {\em idf} factors.
We note that the standard treatment in IR suggests a solution
to this problem that suits batch algorithms --  algorithms that 
determine the weight of a feature after seeing all the examples. We, on the
other hand, seek a solution that can be used in an on-line algorithm. 
Thus, the frequency of a feature throughout the data set, for example, cannot 
be taken into account and we take into account only the {\em tf} term.
We have experimented with three alternative ways of adjusting the value
of $s(f,d)$ according to the frequency of the feature in the document:
(1) Our default is to let the strength indicate only the activity of 
the feature. That is, $s(f,d)=1$, if the feature is present in the document 
(active feature) and $s(f,d)=0$ otherwise.
(2) $s(f,d)=n(f,d)$, where $n(f,d)$ is the number of occurrences of $f$
in $d$; and (3) $s(f,d)=\sqrt{n(f,d)}$ (as in \cite{Wieneretal95}).
These three alternatives examine the tradeoff between the positive and
negative impacts of assigning a strength in proportion to feature
frequency. In most of our experiments, on different data sets, the
choice of using $\sqrt{n(f,d)}$ performed best. 
The results of the comparative evaluation appear in columns 3, 4, and
5 of Table~\ref{tab:our}, corresponding to the three alternatives above.

\subsection{Discarding features}

Multiplicative update algorithm are known to tolerate 
a very large number of features. However, it seems plausible that most 
categories depend only on fairly small subsets of indicative features and 
not on all the features that occur in documents that belong to this class. 
Efficiency reasons, as well as the occasional need to generate 
comprehensible explanations to the classifications, suggest that 
discarding irrelevant features is a desirable goal in IR applications.
If done correctly, discarding irrelevant features may also improve the 
accuracy of the classifier, since irrelevant features contribute noise 
to the classification score. 

An important property of the algorithms investigated here is that they do 
not require a feature selection pre-processing stage. Instead, they can run 
in the presence of a large number of features, and allow for discarding 
features ``on the fly'',  based on their contribution to an accurate 
classification.
This property is especially important if one is considering enriching the 
set of features, as is done in \cite{GoldingRo96,CohenSi96}; in these cases
it is important to allow the algorithm to decide for itself which of the 
features contribute to the accuracy of the classification.

We filter features that are irrelevant for the category based on the
weights they were assigned in the first few training rounds. 

The algorithm is given as input a range of weight value which we call
the {\em filtering range}. 
First, the training algorithm is run for several iterations, until the number
of mistakes on the training data drops below a certain threshold.
After this initial training, we filter out all the features 
whose weight lie in this filtering range. Training then continues as usual.

There are various ways to determine the filtering range. 
The obvious one may be to filter out all features whose weight 
is very close to $0$, but there are a few subtle issues 
involved due to the normalization done in the \PW~algorithm.
In the results presented here we have used, instead, a different filtering range:
Our filtering range is centered around the initial value assigned to 
the weights (as specified earlier for each algorithm), and is bounded 
above and below by the values obtained after one promotion or demotion 
step relative to the initial value. Thus, with high likelihood, we discard
features which have not contributed to many mistakes -- those that were 
promoted or demoted at most once (possibly, with additional promotions and 
demotions which canceled each other, though). 

The results of classification with feature filtering appear in the
last column of Table~\ref{tab:our}. We hypothesize that the
improved results are due to reduction in the
noise introduced by irrelevant features. Further investigation of this
issue will be presented in the long version of this paper.
Typically, about two thirds of the features were filtered for each category, 
significantly reducing the output representation size.

\begin{mytable}[tbh]
\begingroup\catcode`?=\active%
\def?{\kern\digitwidth}%
\begin{center}
\begin{tabular}{||l|c|c||}
\hline
Algorithm & Apte's split & Lewis's split  \\
\hline
\hline
{\bf \BWp}                                 & {\bf 83.3 }       & {\bf 74.7 }   \\
\hline
Experts unigram \cite{CohenSi96}           & 64.7       & 65.6    \\
\hline
Neural Network \cite{Wieneretal95}         & 77.5       & NA       \\
\hline
Rocchio \cite{Rocchio71}                   & 74.5       & 66.0       \\
\hline
Ripper \cite{CohenSi96}                    & 79.6       & 71.9       \\
\hline
Decision trees  \cite{LewisRi94}           & NA         & 67.0       \\
\hline
Bayes  \cite{LewisRi94}                    & NA         & 65.0       \\
\hline
SWAP \cite{Apte-et-al94}                   & 78.9       & NA       \\
\hline
\end{tabular}
\end{center}
\caption{Break-even points comparison.
The data is split into training set and test set based on Lewis's split 
-- \cite{Lewis92}, 14704 documents for training, 6746 for testing, and
Apte's split -- \cite{Apte-et-al94}, 10645 training, 3672 testing, omitting
documents with no topical category.
}
\label{tab:final}
\endgroup
\end{mytable}
\section{Summary of Experimental Results} \label{sec:exp}

The study described in Section~\ref{sec:exp1} was used to determined
the version that performs best, out of those we have experimented with.
Eventually, we have selected the version of the \BW~algorithm, which 
incorporates the $\theta$-range modification, a square-root of occurrences as 
the feature strength and the discard features modification 
(\BWp~in Table~\ref{tab:final}).

We have compared this version with a few other algorithms which have appeared
in the literature on the complete Reuters corpus.
Table~\ref{tab:final} presents break-even points for \BWp~and the other
algorithms, as defined in Section~\ref{sec:methodology}.

The results are reported for two splits of the complete Reuters corpus as
explained in Section 2.3. The
algorithm was run with iterations, threshold range, feature filtering, and
frequency-square-root feature strength. 

The first two rows in Table~\ref{tab:final} compare the 
performance of \BWp~with the two algorithms that most
resemble our approach, the Experts algorithm from \cite{CohenSi96}
and a neural network approach presented in \cite{Wieneretal95}.
(see Section~\ref{sec:related}).
 
Rocchio's algorithm is one of the classical algorithms for this tasks, 
and it still performs very good compared to newly developed techniques 
(e.g, \cite{Lewisetal96}). We also compared with the Ripper algorithm
presented in\cite{CohenSi96} (we present the best results for this task,
with negative tests), a simple decision tree learning system and a 
Bayesian classifier. The last two figure are taken from \cite{LewisRi94} where they
were evaluated only on Lewis's split. The last comparison is with the learning system used
by \cite{Apte-et-al94}, SWAP, which was evaluated only on Apte's split.

Our results significantly outperform (by at least 2-4\%) all results which 
appear in that table and use the same set of features (based on single words).
Of the results we know of in the literature, only a version of the Experts
algorithm of \cite{CohenSi96} which uses a richer feature set -- sparse word 
trigrams --  outperforms our result on the Lewis split, with a
break-even point of 75.3\%, compared with 74.6\% for the unigram-based 
\BWp. However, this version achieves only 75.9\% on the Apte split 
(compared with 83.3\% of \BWp). 
In the long version of this paper we plan to present the results of our algorithm
on a richer feature set as well.

\section{Conclusions} \label{sec:concl}

Theoretical analyses of the Winnow family of algorithms
have predicted an exceptional ability to deal with large numbers 
of features and to adapt to new trends not seen during training.
Until recently, these properties have remained largely
undemonstrated.

We have shown that while these algorithms have many advantages
there is still a lot of room to explore when applying them to a 
real-world problem. 
In particular, we have demonstrated (1) how variation in document length 
can be tolerated through either normalization or negative weights, (2) the
positive effect of applying a threshold range in training, (3)
alternatives in considering feature frequency, and (4) the benefits of
discarding irrelevant features as part of the training algorithm.
The main contribution of this work, however, is that we have presented
an algorithm, \BWp, which performs significantly better than any other 
algorithm tested on these tasks using unigram features. 

We have exhibited that, as expected, multiplicative-update algorithms have 
exceptionally good behavior in high dimensional feature spaces, even in the 
presence of irrelevant features.
One advantage this important property has is that is allows one to decompose 
the learning problem from the feature selection problem. 
Using this family of algorithms frees the designer from the need to 
choose the appropriate set of features ahead of time: A large set of
features can be used and the algorithm will eventually discard those that 
do not contribute to the accuracy of the classifier.
While we have chosen in this study to use a fairly simple set of features, 
it is straight forward to plug in instead a richer set of features. 
We expect that this will further improve the results of the algorithm, although
further research is needed on policies of discarding features and avoidance
of over-fitting. 
In conclusion, we suggest that the demonstrated advantages of the 
Winnow-family of algorithms make it an appealing candidate for further use 
in this domain.

\subsection*{Acknowledgments}
Thank to Michal Landau for her help in running the experiments.

\end{document}